# Prediction of Customer Churn in Banking Industry

Sina E. Charandabi, *Department of Decision Science, LeBow College of Business*, Drexel University

Se456@drexel.edu

## Abstract

With the growing competition in banking industry, banks are required to follow customer retention strategies while they are trying to increase their market share by acquiring new customers. This study compares the performance of six supervised classification techniques to suggest an efficient model to predict customer churn in banking industry, given 10 demographic and personal attributes from 10000 customers of European banks. The effect of feature selection, class imbalance, and outliers will be discussed for ANN and random forest as the two competing models. As shown, unlike random forest, ANN doesn't reveal any serious concern regarding overfitting and is also robust to noise. Therefore, ANN structure with five nodes in a single hidden layer is recognized as the best performing classifier.

## Introduction

With the growing competition in banking industry, banks are required to follow customer retention strategies while they are trying to increase their market share by acquiring new customers. It is shown that improving the retention rate by up to 5 % can increase a bank's profit up to 85 % (Nie et al., 2011). Additionally, attracting new customers costs more to any company rather than retaining the old ones who are likely to produce more profit (Verbeke et al., 2011). Thus, banks should maintain their competitive advantage by taking the advantage of machine learning models to predict customer churn. Given a randomly sampled population of 10000 customers from three European-based banks, this project intends to propose an efficient predictive model for customer churn in banking industry, using different supervised classification techniques. Model performance, goodness of fit, feature selection, class imbalance and dealing with outliers will be discussed in the following sections.

## Data Pre-Processing

Dataset selected for this study is publically available in kaggle.com[1]. Variables included in the dataset are described in Table 1. Out of 13 variables, CustomerId and Surname need to be removed as they don't have any contribution to the classification purpose. We also replace binary values of the outcome variable (Exited) with "Stayed" and "Left" labels to have a better representation of outputs when visualizing results and discussing the performance. We will use data entirely in the analysis and don't follow any sampling procedure because we need the training sample to be sufficiently large.

---

[1] https://www.kaggle.com/barelydedicated/bank-customer-churn-modeling

Current data doesn't have any missing value in none of its 10000 observations and thus, there won't be any concern in this regard. However, customers who stayed with banks (7963 customers) are around four times the number of those who left (2037 customers). Therefore, data is imbalance with respect to the outcome variable and this concern needs to be addressed in the modeling section.

We also need to figure out potential outliers in at each class of the outcome variable for all numeric variables. As depicted in Figures 1 to 4, Balance and Estimated Salary don't include any outliers. Credit Score and Age have only 11 and 13outliers, respectively, in the "Left" class and Age includes 486 outliers in the "Stayed" class. Therefore, in general, there is not a serious concern with regard to outliers as the ratio of outliers-where they were detected- to the size of data is reasonably low. However, we will analyze data in the absence of these 486 outliers - associated with Age (still consisting only 0.05% of data) - to address any possible noise from these data points in the evaluation of the final model.

Table 1. Variable Definitions and Descriptive Statistics

| Variable | Type | Definition | Minimum | Maximum | Mean | Std. Deviation |
|---|---|---|---|---|---|---|
| CustomerId | Nominal | Customer ID | | | | |
| Surname | Text | | | | | |
| CreditScore | Interval | Customer's credit score | 350 | 850 | 650.53 | 96.65 |
| Geography | Nominal | France, Germany, Spain | | | | |
| Gender | Nominal | Female, Male | | | | |
| Age | Ratio | | 18 | 92 | 38.92 | 10.49 |
| Tenure | Interval | Tenure of deposit | 0 | 10 | 5.01 | 2.89 |
| Balance | Ratio | | 0 | 250898.1 | 76485.89 | 62397.41 |
| NumOfProducts | Interval | Number of bank account affiliated products the customer has | 1 | 4 | 1.53 | 0.58 |
| HasCrCard | Binary | Does the customer have a credit card through the bank? (Yes=1, No=0) | 0 | 1 | 0.71 | 0.46 |
| IsActiveMember | Binary | Is the customer an active member? (Yes=1, No=0) | 0 | 1 | 0.52 | 0.50 |
| EstimatedSalary | Ratio | Estimated salary of the customer | 11.58 | 199992.5 | 100090.2 | 57510.49 |
| Exited | Binary | Did the customer leave the bank within the last 6 month? (Yes=1, No=0) | 0 | 1 | 0.2 | 0.40 |

\* N=10000, This Dataset lacks any missing value

\* Out of 13 columns listed in Table 1, CustomerId and Surname will be excluded from the analysis.

Lastly, we need to explore the distribution of data across numeric variables to see whether we may expect any necessary data transformation step prior to the modeling section. As depicted in Figure 1 to 4, given the majority of customers with credit score between 600 and 700, the distribution is fairly normal in this respect. Given majority of customers at around 40 year old, age is also fairly normal with a slight skewness to right. However, majority of clients have '0' balance while that of the remaining ones follows a decent normal distribution. The distribution of estimated salary is also pretty uniform and obviously not normal. As Tenure and Number of products include only few integer values, we use bar plot instead of histogram to discuss their distribution. Tenure's distribution is symmetric with the lowest values at each end, associating with deposits tenured in less than a year or 10 years (Figure 5). On the other hand, the distribution of Number of Products isn't symmetric, having the largest number of customers subscribed to one or two products at most and less than 500 customers associated with more products (Figure 6). Therefore, data transformation is recommended when normality assumption needs to be satisfied.

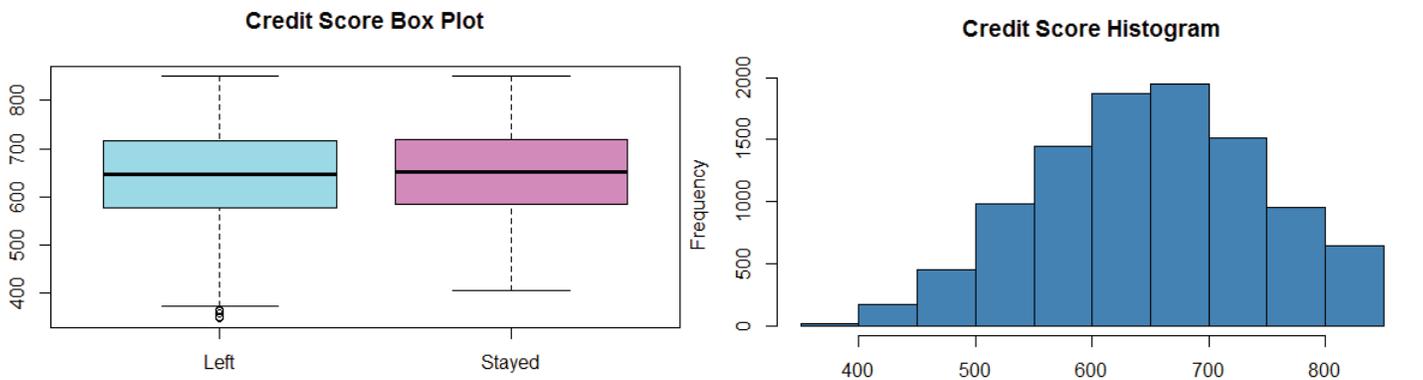

Figure 1. Credit Score Distribution

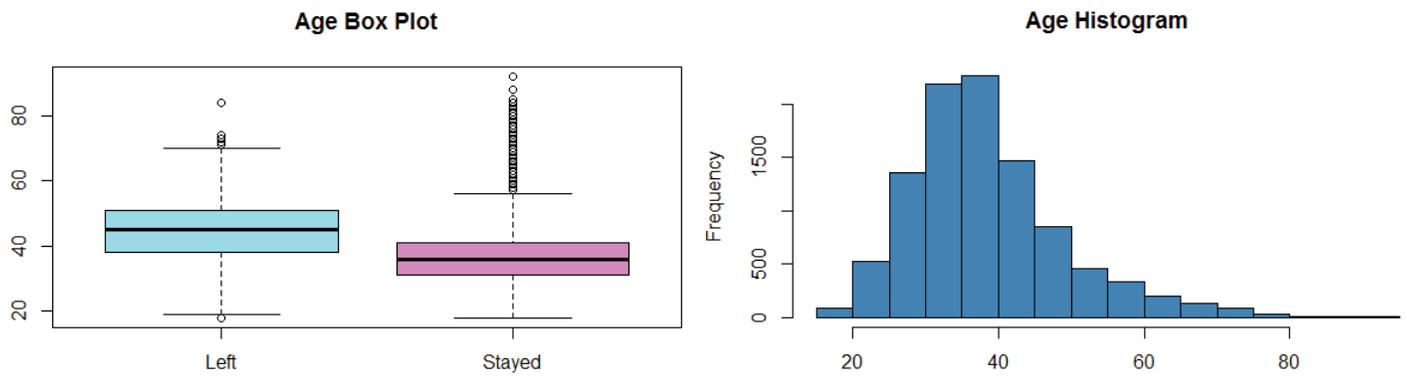

Figure 2. Age Distribution

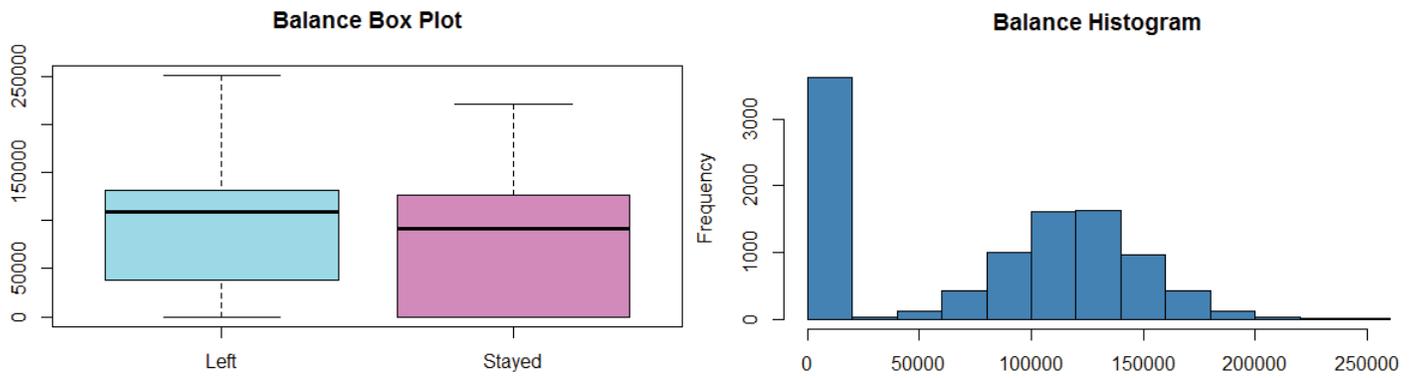

Figure 3. Balance Distribution

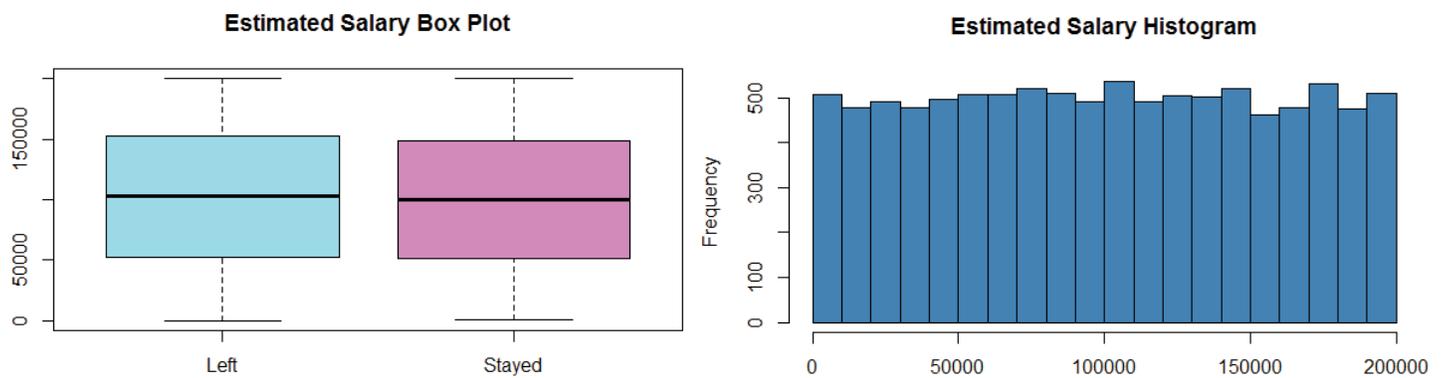

Figure 4. Estimated Salary Distribution

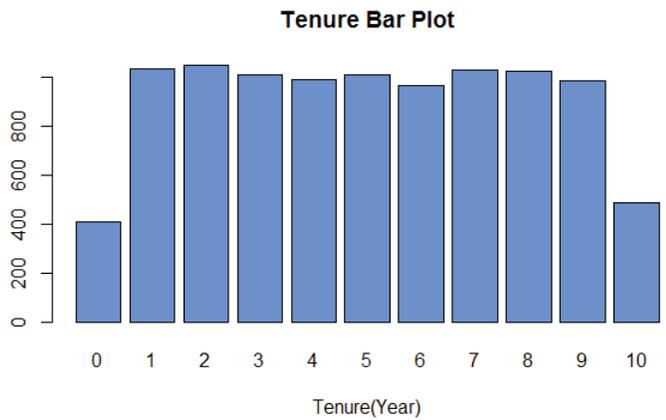

Figure 5. Tenure Distribution

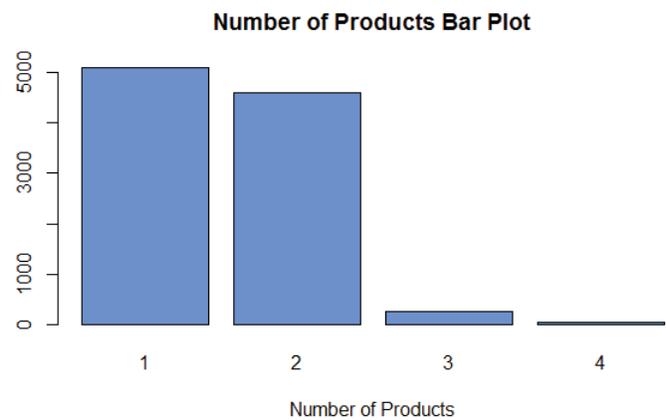

Figure 6. Number of Products Distribution

As shown in Table 2, none of independent variables (IVs) are strongly correlated with each other and Balance and Number of Products uncover the highest correlation (-0.304). Therefore, there shouldn't be any concern regarding the existence of redundant variables in the primary model. None of IVs are also strongly correlated with the DV, given Age and IsActiveMember with the two highest values (0.256, -0.156, respectively).

Table 2. Correlation Table

| | Exited | CreditScore | Age | Tenure | Balance | NumOfProducts | EstimatedSalary | IsActiveMember | HasCrCard |
|---:|---|---|---|---|---|---|---|---|---|
| **Exited** | 1 | | | | | | | | |
| **CreditScore** | -0.027 | 1.000 | | | | | | | |
| **Age** | 0.285 | -0.004 | 1.000 | | | | | | |
| **Tenure** | -0.014 | 0.001 | -0.010 | 1.000 | | | | | |
| **Balance** | 0.119 | 0.006 | 0.028 | -0.012 | 1.000 | | | | |
| **NumOfProducts** | -0.048 | 0.012 | -0.031 | 0.013 | -0.304 | 1.000 | | | |
| **EstimatedSalary** | 0.012 | -0.001 | -0.007 | 0.008 | 0.013 | 0.014 | 1.000 | | |
| **IsActiveMember** | -0.156 | 0.026 | 0.085 | -0.028 | -0.010 | 0.010 | -0.011 | 1.000 | |
| **HasCrCard** | -0.007 | -0.005 | -0.012 | 0.023 | -0.015 | 0.003 | -0.010 | -0.012 | 1.000 |

## Exploratory Data Analysis

In order to explore the dependency between churning behavior and classes of categorical variable included in the model, we visualize the distribution of two types of customers - as the dependent variable (DV) - at each class and also perform Chi-Square test to detect significant dependencies. Data is pretty balanced across males and females (Figure 7) and females represent significantly lower churning rate than males. (X-squared = 112.92, df = 1, p-value < 2.2e-16). Data is also pretty balanced between banks from Germany and Spain while records from France-based banks are in majority, around twice those of other two banks (Figure 8) and customers of French banks associate with a significantly lower churning rate (X-squared = 301.26, df = 2, p-value < 2.2e-16). Having credit card from a bank (Figure 9) doesn't correspond with a significant difference in its churning rate (X-squared = 0.47134, df = 1, p-value = 0.4924) while active customers associate (Figure 10) with a higher retention rate rather than non-active ones (X-squared = 242.99, df = 1, p-value < 2.2e-16). As indicated in Figure 11[2], customers who purchased one or two products stay with the bank for the most part. On the other hand, most of those who purchased three or four products left the bank (X-squared = 1503.6, df = 3, p-value < 2.2e-16).

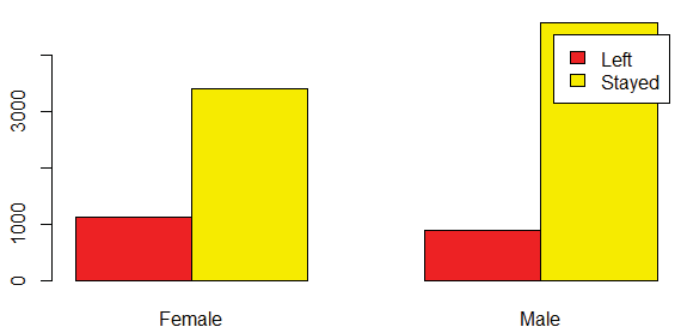

Figure 7. Churning Behavior across gender

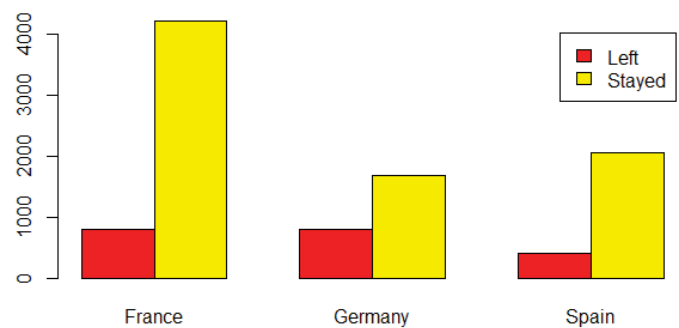

Figure 8. Churning Behavior across countries

---

[2] Although we dealt with NumOfProducts as a numeric variable, we can still study the distribution of outcome variable across its only 4 values.

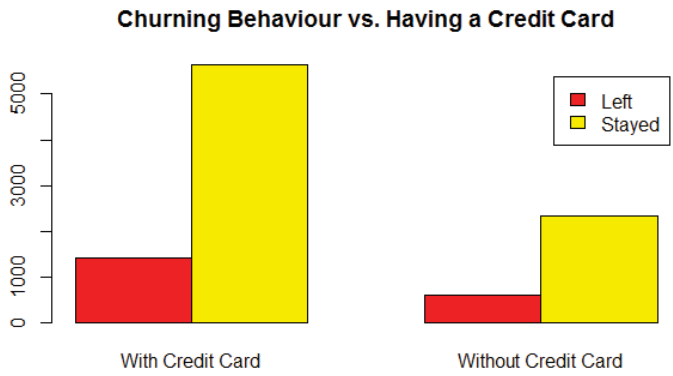

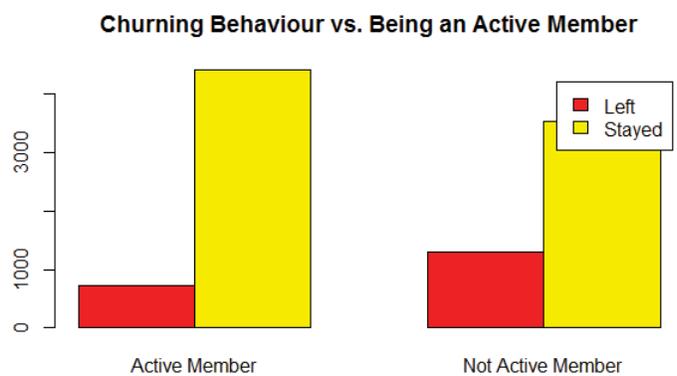

Figure 9. Churning vs. having a credit card

Figure 10. Churning vs. being an active member

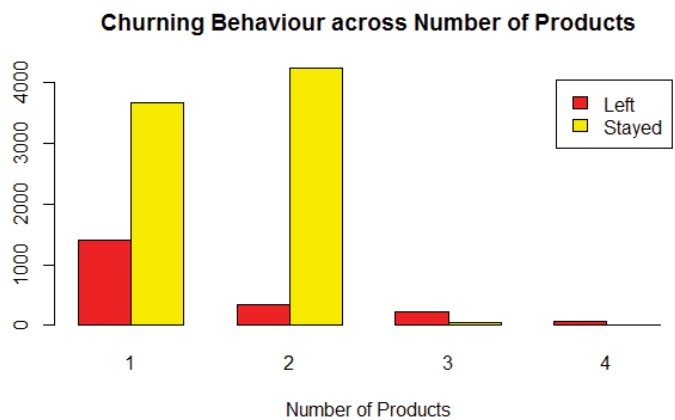

Figure 11. Churning behavior across the number of products

Thus, we can only deduce that there is dependency between churning behavior and the number of accounts a customer is affiliated with (Notice that the churn ratio doesn't represent a constantly decreasing or increasing trend with respect to the number of products.)

## Analysis

### Model Selection

Regarding the binary nature of the outcome variable, current problem is well-suited with almost all supervised classification algorithms and tree-based models appear to be the popular ones as suggested by the literature (Nie et al., 2011, Verbeke et al., 2011). Reminding that our data lacks missing values and redundant features; and also includes not a large number of IVs which result in irrelevant features, computational costs and overfitting issues are likely to determine why some models might be preferred to others. To this end, we first make a comparison between the performances of different classification techniques in order to candidate two competing models for further analysis.

Naïve Bayes is the only technique which requires a data transformation step to ensure that numerical IVs follow Gaussian distribution to a reasonable extent. However, while the effect of transformation is visible for the estimated salary through making it more inclined to normal rather than uniform, Balance distribution still suffers from the presence of the largest number of customers with zero balance.

Due to the existence of categorical variables (Gender and Geography), their associated dummies are required to be created prior to the modeling by K-NN, SVM and ANN. Additionally, when implementing SVM and ANN, z-scores rescaling and min-max normalization over 0-1 range also need to be applied, respectively, in order to return accurate results. On the other hand, when employing decision trees and random forest, none of these further pre-processing steps are required.

Hyperparameter tuning is another step which makes the modeling step by k-NN, decision trees, random forest and ANN time-consuming. To do so, using a grid search with accuracy as the performance measure and a repeated 10-Fold cross validation for three times, we achieved the optimal number of nearest neighbors (k=9), complexity parameter (cp=0.01) for decision trees, the optimal number of variables to randomly sample as potential variables to split on (mtry=4) for random forest, and the optimal number of the hidden layer's nodes (size=5) and weight decay (0.1) for ANN. While tuning procedure converged to the optimal solution within 10 to 15 minutes for aforementioned techniques, SVM turned out to be time demanding for a greater extent, and inevitably we provided the result for its not-tuned model.

Table 3 and 4 show the performance measures for all developed models after splitting the data into training and testing sets using an 80/20 split rule. As suggested by testing models over the unseen data, for Naïve Bayes and k-nn, Kappa is neither poor nor good but fair (0.2 < kappa < 0.4) and F1 is also closer to 0 rather than 1. Thus, the overall performance suggested by these models is not good. On the other hand, it is seen that SVM, even without tuning its hyperparameters performs better than both of these models. For SVM and decision trees, Kappa achieves a larger value which is not still good but moderate (0.4 < kappa < 0.5) and F1 gets closer to 1 rather than 0, hence resulting in a moderate performance. Following the implementation of random Forest and ANN, and achieving larger values of kappa (0.5 < kappa < 0.6) and F1 the model improves to some extent. As appears, overfitting is substantial for random forest while models developed upon Naïve Bayes and decision trees are slightly underfitted.

Based on our findings, Naïve Bayes and particularly k-nn have the poorest performance and thus being discarded from further analysis. As expected, the violation of normality assumption seems to negatively affect the performance of Naïve Bayes model. SVM is also discarded mainly due to its slow speed in training phase and the selecting the optimal parameters, though being perfectly robust to overfitting. From the remaining three models, ANN would definitely be the first candidate because it not only suggests the overall best values for classification metrics, but it is also robust to overfitting. Finally, we prefer to select random forest as the second candidate. Although being evidently prone to overfitting, it still performs slightly better than decision trees in general and specifically, feature selection and balancing data are likely to address its weaknesses to some extent.

Table 3. Performance measures for the training set

|  | Naïve Bayes | k-nn | SVM | Decision Trees | Random Forest | ANN |
|---|---|---|---|---|---|---|
| Kappa | 0.274 | 0.381 | 0.489 | 0.463 | 1.000 | 0.513 |
| Accuracy | 0.815 | 0.841 | 0.865 | 0.858 | 1.000 | 0.864 |
| Precision | 0.609 | 0.756 | 0.845 | 0.802 | 1.000 | 0.762 |
| Recall | 0.258 | 0.328 | 0.417 | 0.402 | 1.000 | 0.481 |
| F1 | 0.362 | 0.457 | 0.558 | 0.536 | 1.000 | 0.590 |

Table 4. Performance measures for the testing set

|  | Naïve Bayes | k-nn | SVM | Decision Trees | Random Forest | ANN |
|---|---|---|---|---|---|---|
| Kappa | 0.310 | 0.257 | 0.488 | 0.492 | 0.512 | 0.514 |
| Accuracy | 0.824 | 0.817 | 0.866 | 0.865 | 0.864 | 0.867 |
| Precision | 0.663 | 0.648 | 0.853 | 0.835 | 0.772 | 0.797 |
| Recall | 0.280 | 0.226 | 0.413 | 0.423 | 0.474 | 0.464 |
| F1 | 0.394 | 0.335 | 0.556 | 0.561 | 0.588 | 0.587 |

## Feature Selection

First, we use embedded methods in order to rank IVs based on their discriminating power between two classes. To do so, variable importance outputs suggested by both decision trees and random forest models are taken into account. As shown in Table 5, compared to decision trees, the rank of variables used in random forest is significantly different in terms of their importance. Specifically, variables of tenure, credit score and estimated salary which appear to be less important by decision trees, turned out to be more impactful by random forest. On the other hand, being located in Germany is recognized less important by random forest while it is suggested as an important feature relying on decision trees. Additionally, random forest itself doesn't suggest a consistent ranking structure depending on whether accuracy or Gini is selected for comparing between mean decrease by the exclusion of input variables (Figure 12). Accordingly, we can only say that having credit card is the least important variable by either criterion and age along with balance is among the top four important features. However, discarding HasCard variable and keeping only Age and Balance as they are the only two consistent variables by mean decrease criteria isn't a meaningful decision for this problem. Thereby, we use recursive feature elimination (RFE) from wrapper methods in order to obtain a more sound and persuasive basis for feature selection.

Table 5. Variable importance by tree-based models

| Decision Trees | | Random Forest | |
| --- | --- | --- | --- |
| **Variable** | **Score** | **Variable** | **Score** |
| **NumOfProducts** | 100 | **Age** | 100 |
| **Age** | 85.858 | **Balance** | 58.268 |
| **GeographyGermany** | 40.732 | **EstimatedSalary** | 57.314 |
| **IsActiveMember** | 33.616 | **CreditScore** | 55.235 |
| **Balance** | 21.253 | **NumOfProducts** | 49.417 |
| **GenderMale** | 6.007 | **Tenure** | 28.345 |
| **EstimatedSalary** | 1.906 | **IsActiveMember** | 12.607 |
| **Tenure** | 1.545 | **GeographyGermany** | 6.363 |
| **CreditScore** | 0 | **GenderMale** | 2.433 |
| **HasCrCard** | 0 | **HasCrCard** | 1.903 |
| **GeographySpain** | 0 | **GeographySpain** | 0 |

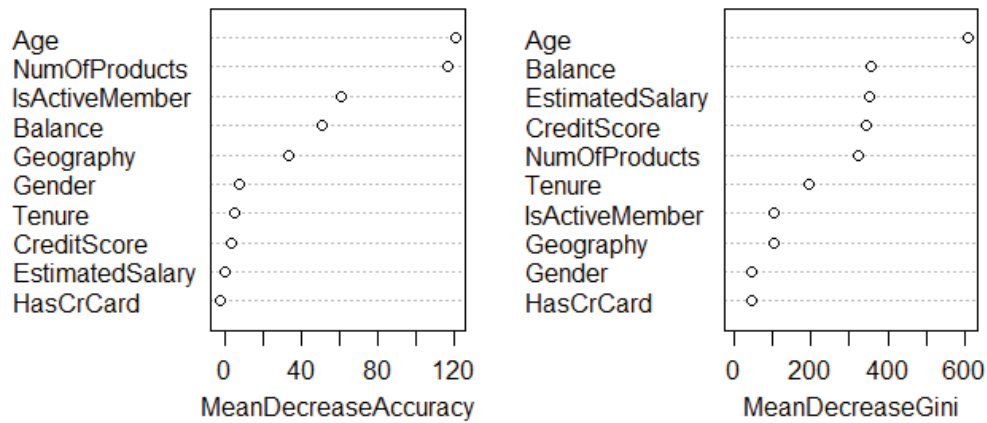

Figure 12. Variable importance by mean decrease criterion (random forest)

RFE suggests the inclusion of all 10 variables (accuracy=0.8628, Kappa= 0.5081) which will result in what we have obtained so far. Thus, in order to test the model under a different condition, we select only the top five features which are the same top five variables detected by both decision trees and mean decrease in accuracy by random forest. These variables include: Age, NumOfProducts, IsActiveMember, Balance, Geography. It is worth mentioning that given only four variables, random forest still performs very closely to the initial model (accuracy=0.8503, Kappa= 0.4441). Therefore, restricting variables to the top five features seems to be a reasonable decision.

In Table 6 and 7 the performance of random forest and ANN models was compared prior to and after feature selection. We used the same split ratio (80/20) and hyperparameter tuning specifications in this section. ANN is converged to the same optimal size (5) and decay values (0.1) but random forest now requires one more variable to randomly sample as potential variables to split on (mtry=5). It is seen that excluding five least important variables affects the classification result at a really slight degree (positively for ANN but negatively for random forest). Thus, we can have a simpler and less-complicated model while obtaining the same predictions. By the way, overfitting is still the main concern with regard to random forest.

Table 6. Performance measures following feature selection for the training set

|  | Random Forest | | ANN | |
| --- | --- | --- | --- | --- |
|  | Initial | After Feature Selection | Initial | After Feature Selection |
| Sensitivity | 1.000 | 0.571 | 0.481 | 0.502 |
| Specificity | 1.000 | 0.978 | 0.962 | 0.961 |
| Kappa | 1.000 | 0.895 | 0.513 | 0.531 |
| Accuracy | 1.000 | 0.628 | 0.864 | 0.867 |
| Precision | 1.000 | 0.868 | 0.762 | 0.767 |
| Recall | 1.000 | 0.571 | 0.481 | 0.502 |
| F1 | 1.000 | 0.688 | 0.590 | 0.607 |

Table 7. Performance measures following feature selection for the testing set

|  | Random Forest | | ANN | |
| --- | --- | --- | --- | --- |
|  | Initial | After Feature Selection | Initial | After Feature Selection |
| Sensitivity | 0.474 | 0.464 | 0.464 | 0.479 |
| Specificity | 0.964 | 0.962 | 0.970 | 0.965 |
| Kappa | 0.512 | 0.499 | 0.514 | 0.519 |
| Accuracy | 0.864 | 0.861 | 0.867 | 0.866 |
| Precision | 0.772 | 0.759 | 0.797 | 0.780 |
| Recall | 0.474 | 0.464 | 0.464 | 0.479 |
| F1 | 0.588 | 0.576 | 0.587 | 0.594 |

**Class Imbalance**

As we noticed, churned customers are in minority, around a quarter of the loyal customers, thus causing class imbalance in the data. To address this issue, we should decide whether to use under-sampling or oversampling techniques. The advantage of under-sampling over the "Stayed" class is that we are not creating any artificial data points and because the original data is sufficiently large we are not concerned about missing too much information. However, because the gap between two classes is not too large, oversampling of the "Left" class doesn't also lead to serious concerns regarding the insertion of artificially-created data points. Thereby, both techniques seem to be appropriate for this problem. In this section we will perform both resampling types over the recently-created training and testing sets which include only the top five attributes. SMOTE technique will be used for oversampling purpose.

Again, we are using the same split ratio and hyperparameter tuning specifications. ANN is converged to the same optimal size (5) but an increased decay value (0.2) when SMOTE is applied. This time, random forest requires only two variable to randomly sample as potential variables to split on the under-sampled data (mtry=2) while this number is increased to five when the data is up-sampled using SMOTE.

As indicated in Table 9, overall improvement through balancing data in the random forest model is substantial with respect to all measures except for recall and under-sampling improves performance measures other than precision to a greater extent than SMOTE. On the other hand, in ANN model the overall improvement through balancing data is not substantial with respect to all measures except for recall. Under-sampling and SMOTE, either one, improves the performance measures to a different extent and in terms of Precision and Recall, SMOTE suggests a bit better results while for F1 and Accuracy, under-sampling performs better.

SMOTE also returns a larger ROC value than under-sampling and random forest achieves the largest ROC value through it. However, given the performance metrics from the training set (Table 8) random forest still seems more prone to suggest an overfitted model, particularly in SMOTE mode while ANN model results in overfitting to a lower extent, particularly in SMOTE mode.

Given these results, ANN model oversampled by SMOTE method seems to be a slightly better predictive model than random forest for this problem.

**Table 8. Performance measures following balancing data for the training set**

|  | Random Forest | | | ANN | | |
|---|---|---|---|---|---|---|
|  | Initial* | Under-Sampled | Over-Sampled | Initial* | Under-Sampled | Over-Sampled |
| Sensitivity | 1.000 | 0.572 | 0.967 | 0.481 | 0.766 | 0.714 |
| Specificity | 1.000 | 0.979 | 0.991 | 0.962 | 0.819 | 0.849 |
| Kappa | 1.000 | 0.551 | 0.961 | 0.513 | 0.585 | 0.569 |
| Accuracy | 1.000 | 0.775 | 0.981 | 0.864 | 0.793 | 0.791 |
| Precision | 1.000 | 0.965 | 0.988 | 0.762 | 0.809 | 0.780 |
| Recall | 1.000 | 0.572 | 0.967 | 0.481 | 0.766 | 0.714 |
| F1 | 1.000 | 0.718 | 0.977 | 0.590 | 0.787 | 0.745 |
| ROC | – | 0.855 | 0.936 | – | 0.861 | 0.870 |

* Initial models are using only selected features.

**Table 9. Performance measures following balancing data for the testing set**

| Random Forest | ANN |
|---|---|

|             | Initial* | Under-Sampled | Over-Sampled | Initial* | Under-Sampled | Over-Sampled |
|---|---|---|---|---|---|---|
| Sensitivity | 0.474 | 0.781 | 0.631 | 0.464 | 0.764 | 0.698 |
| Specificity | 0.964 | 0.803 | 0.856 | 0.970 | 0.809 | 0.861 |
| Kappa | 0.512 | 0.484 | 0.454 | 0.514 | 0.482 | 0.512 |
| Accuracy | 0.864 | 0.798 | 0.810 | 0.867 | 0.714 | 0.828 |
| Precision | 0.772 | 0.503 | 0.528 | 0.797 | 0.506 | 0.561 |
| Recall | 0.474 | 0.781 | 0.631 | 0.464 | 0.764 | 0.698 |
| F1 | 0.588 | 0.612 | 0.575 | 0.587 | 0.609 | 0.622 |

* Initial models are using only selected features.

## Effect of Outliers

As discussed earlier, presence of few outliers in Age variable which correspond with the "Stayed" class of DV are likely to cause some noise in the model, although it would be at a really slight degree, given the larger size of the data. In this section, we clean data from outliers and re-create models this time for selected features and data oversampled by SMOTE Method. Splitting data and hyperparameter tuning procedures stay the same.

Table 10. Performance measures in the absence of outliers for the training set

|  | Random Forest | | ANN | |
|---|---|---|---|---|
|  | Initial* | Cleaned | Initial* | Cleaned |
| Sensitivity | 0.967 | 0.973 | 0.714 | 0.733 |
| Specificity | 0.991 | 0.989 | 0.849 | 0.852 |
| Kappa | 0.961 | 0.963 | 0.569 | 0.560 |
| Accuracy | 0.981 | 0.982 | 0.791 | 0.801 |
| Precision | 0.988 | 0.984 | 0.780 | 0.787 |
| Recall | 0.967 | 0.973 | 0.714 | 0.733 |
| F1 | 0.977 | 0.979 | 0.745 | 0.759 |

* Initial models are using only selected features and oversampled data with SMOTE.

Table 11. Performance measures in the absence of outliers for the testing set

|  | Random Forest | | ANN | |
|---|---|---|---|---|
|  | Initial* | Cleaned | Initial* | Cleaned |
| Sensitivity | 0.631 | 0.581 | 0.698 | 0.732 |
| Specificity | 0.856 | 0.829 | 0.861 | 0.833 |
| Kappa | 0.454 | 0.369 | 0.512 | 0.485 |
| Accuracy | 0.810 | 0.781 | 0.828 | 0.813 |
| Precision | 0.528 | 0.449 | 0.561 | 0.512 |
| Recall | 0.631 | 0.581 | 0.698 | 0.732 |
| F1 | 0.575 | 0.507 | 0.622 | 0.602 |

* Initial models are using only selected features and oversampled data with SMOTE.

As indicated in Table 10 and 11, excluding outliers doesn't improve the performance and interestingly, the results are slightly worse for random forest after diminishing the size of the majority class following the removal of outliers. Thus, again ANN seems to be more robust to noises of this type.

## Discussion

In this study we compared the performance of different supervised classification techniques to suggest an efficient model to predict customer churn in banking industry, given 10 demographic and personal attributes from 10000 customers of European banks. Among six different models we developed, random forest and ANN appeared to be the superior ones in terms of overall performance; however, random forest turned out to be remarkably vulnerable to overfitting, an issue which wasn't resolved satisfactorily after feature selection, but was minimized after addressing the class imbalance. On the other hand, ANN didn't demonstrate any serious concern regarding overfitting and improved the performance substantially over the balanced data. It was also shown that ANN is robust to outliers while the intervention of such noises in the unseen data can adversely affect the performance of random forest to a greater extent. Therefore, based on our findings, ANN structure with five nodes in a single hidden layer excels other classifiers (Figure A. 1, A. 2) to distinguish churned customers. Accordingly, decision trees can be considered as the second best model because it doesn't reveal overfitting issues which random forest is facing.

One advantage of decision trees is that unlike ANN it enables us to rank variables in terms of their contribution to the classification procedure (Figure A. 2). Depending on the context of application, banks might also be interested in this piece of information, making decision trees a better candidate for the prediction. As we noticed, age, number of products a customer is affiliated with, being an active member, balance and being located in Germany are the most important features to predict the type of customers and we can create a parsimonious model by discarding other attributes without being concerned about weakening the performance. Presence of these variables as the most important features is in accordance with what we noticed about the significant dependency between these variables and churning behavior in the exploratory data analysis step (Figure 7 to 11).

As the greedy approach by decision trees suggests (Figure A. 2), younger customers (under 41) are more likely to stay with banks. This could be due to expectations that more-experienced customers have developed during years of their activity in different banks as opposed to younger customers who are supposed to be less-experienced in this respect. Thus, it seems that older customers need to be in priority when targeting loyalty programs. As revealed by the decision tree, older customers who are active members tend to stay. Therefore, banks need to be more cognizant of those with less activity and engage them by periodic and effective marketing campaigns. Number of accounts affects the churning behavior differently, as we noticed earlier in Figure 11. It could be said that, for the most part, customers with one or two accounts are likely to stay with the bank while opening more accounts may result in churning behavior. Although balance appears to be the least impactful variable to predict the customer churn, balance amounts fewer than 32,000 for customers over 61 year old could be an indication of leaving the bank, another point which banks might be consider when setting their priorities. It is also seen that churning behavior is more prevalent among customers from banks of Germany rather than those in France and Spain (in accordance with Figure 8); however without having supporting information we are unable to discuss this observation from the managerial standpoint.

Another interesting implication by this study is that gender, although being significantly dependent on the churning behavior (Figure 7), doesn't appear to be an important predictor of the customer type, a point that banks might consider when designing marketing campaigns and positioning their products. Likewise, other industry-related attributes, including tenure of deposit, having credit card, credit score and salary level also have less contribution to the prediction of the class membership. In this regard, traditional efforts to promote the bank's brand by proposing attractive short-term or long-term deposits and credit cards aren't assumed to guarantee the customer loyalty in the long run.

The sample used in this study was restricted to only European banks and therefore generalization of our findings requires replication of the study over similar samples from other regions. Additionally, data can be enriched if it is combined with the information from the operations and outcomes of marketing campaigns because maintaining marketing communications is an important feature which encourages a customer to stay connected with the bank. The main limitation with all models discussed in this study, including ANN, decision trees and random forest, is that they are unable to help us draw any inference about the statistical significance of contributing factors. Therefore, current study can be developed by creating a logistic regression model to address this weakness.

## Appendix

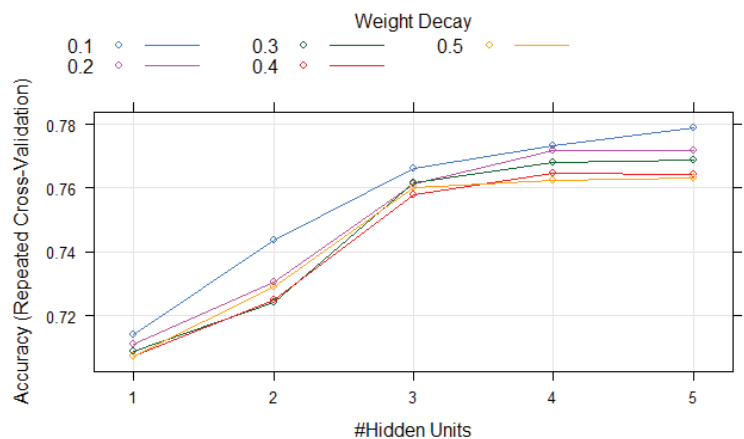

**Figure A. 1. Optimal solution for the ANN structure: under-sampled data with selected features**

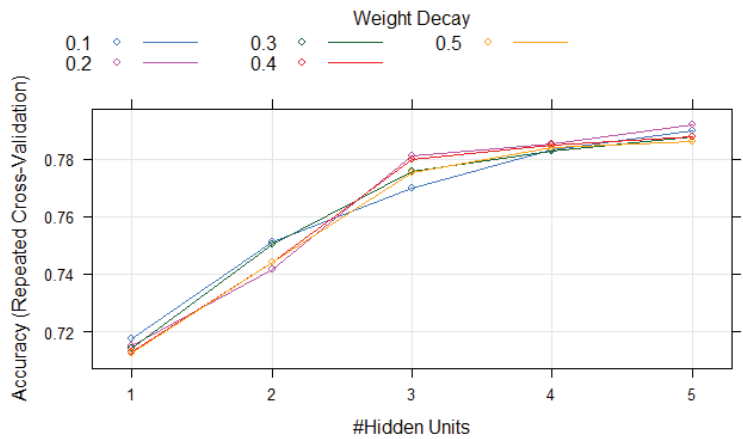

**Figure A. 2. Optimal solution for the ANN structure: over-sampled data with selected features**

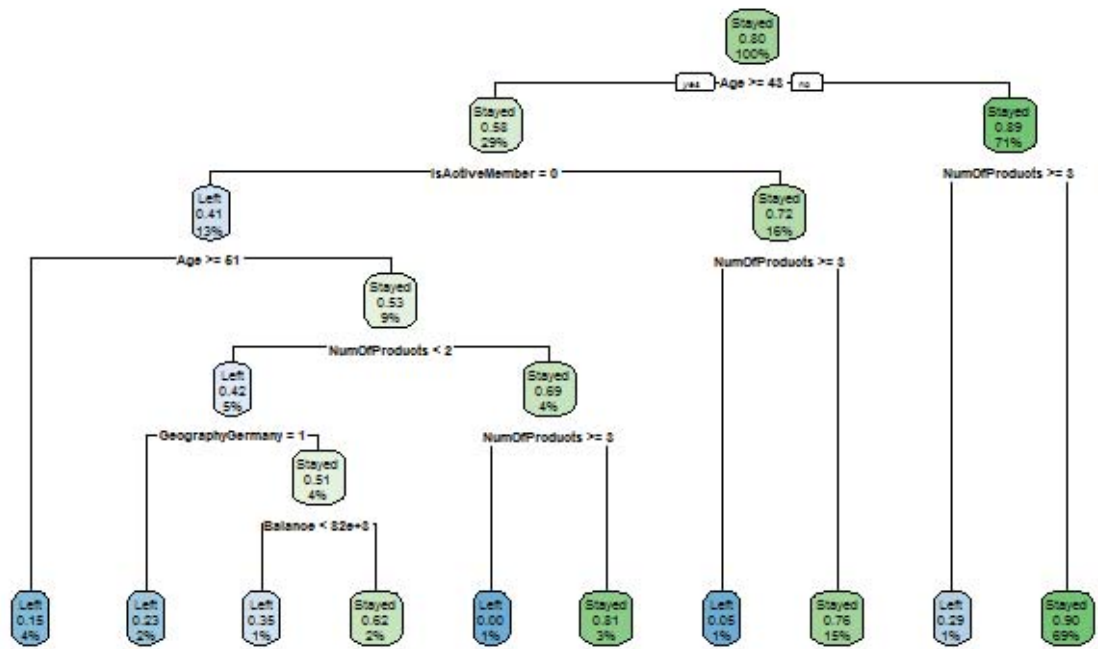

**Figure A. 3. Classification procedure by decision trees approach**